\documentclass{article}

    \PassOptionsToPackage{numbers, compress}{natbib}

\usepackage[final]{neurips_2024}
\usepackage{graphicx}

\usepackage{amsmath}
\usepackage{amssymb}
\usepackage{mathtools}
\usepackage{amsthm}
\usepackage{tcolorbox}
\usepackage{enumitem}
\usepackage{setspace}
\usepackage{pifont}
\usepackage[T1]{fontenc}
\usepackage{graphicx}
\usepackage{caption}

\usepackage[draft,textsize=footnotesize,textwidth=15mm]{todonotes}

\usepackage{amssymb}
\usepackage{pifont}
\newcommand{\cmark}{{\ding{51}}}%
\newcommand{\xmark}{{\ding{55}}}%

\newcommand{\specialcell}[2][c]{%
  \begin{tabular}[#1]{@{}c@{}}#2\end{tabular}}

\usepackage{neurips_2024}

\usepackage[utf8]{inputenc} 
\usepackage[T1]{fontenc}    
\usepackage{hyperref}       
\usepackage{url}            
\usepackage{booktabs}       
\usepackage{amsfonts}       
\usepackage{nicefrac}       
\usepackage{microtype}      
\usepackage{xcolor}         
\usepackage{color, colortbl}

\definecolor{Gray}{gray}{0.9}

\usepackage[capitalize,noabbrev]{cleveref}

\title{Source-Free Domain Adaptation with Diffusion-Guided Source Data Generation}

\author{
  Shivang Chopra \\
  Georgia Institute of Technology\\
  \texttt{shivangchopra11@gatech.edu} \\
  \And
  Suraj Kothawade \\
  University of Texas, Dallas \\
  \texttt{suraj.kothawade@utdallas.edu} \\
  \AND
  Houda Aynaou \\
  Georgia Institute of Technology \\
  \texttt{haynaou3@gatech.edu} \\
  \And
  Aman Chadha\thanks{Work does not relate to position at Amazon.} \\
  Amazon GenAI \\
  \texttt{hi@aman.ai} \\
}

\begin{document}

\maketitle

\begin{abstract}
  This paper introduces a novel approach to leverage the generalizability of \textbf{D}iffusion \textbf{M}odels for \textbf{S}ource-\textbf{F}ree \textbf{D}omain \textbf{A}daptation (\textbf{DM-SFDA}). Our proposed DM-SFDA method involves fine-tuning a pre-trained text-to-image diffusion model to generate source domain images using features from the target images to guide the diffusion process. Specifically, the pre-trained diffusion model is fine-tuned to generate source samples that minimize entropy and maximize confidence for the pre-trained source model. We then use a diffusion model-based image mixup strategy to bridge the domain gap between the source and target domains. We validate our approach through comprehensive experiments across a range of datasets, including Office-31~\cite{office31}, Office-Home~\cite{office_home}, and VisDA~\cite{visda2017}. The results demonstrate significant improvements in SFDA performance, highlighting the potential of diffusion models in generating contextually relevant, domain-specific images.
\end{abstract}

\section{Introduction}
\label{sec:intro}
Deep Convolutional Neural Networks (CNNs) have demonstrated impressive performance on several visual tasks in recent years. However, the assumption that the distributions of the training and test sets are the same is crucial to the effectiveness of CNNs~\cite{sfda-de}. Subsequently, a big drop in performance is generally observed when CNN-based models are deployed in real-world settings with a discrepancy in data distribution~\cite{usfda}. Domain Adaptation (DA) attempts to reduce this disparity to make these models perform well across multiple domains. Traditional DA approaches that rely on fixed source data might struggle to keep up with the pace of domain changes. Moreover, the rising prominence of data privacy regulations has led to a demand for DA techniques that can function effectively without relying on access to the source training data, a setting generally known as Source Free Domain Adaptation (SFDA). 


\begin{figure*}
    \centering
    \includegraphics[scale=0.4]{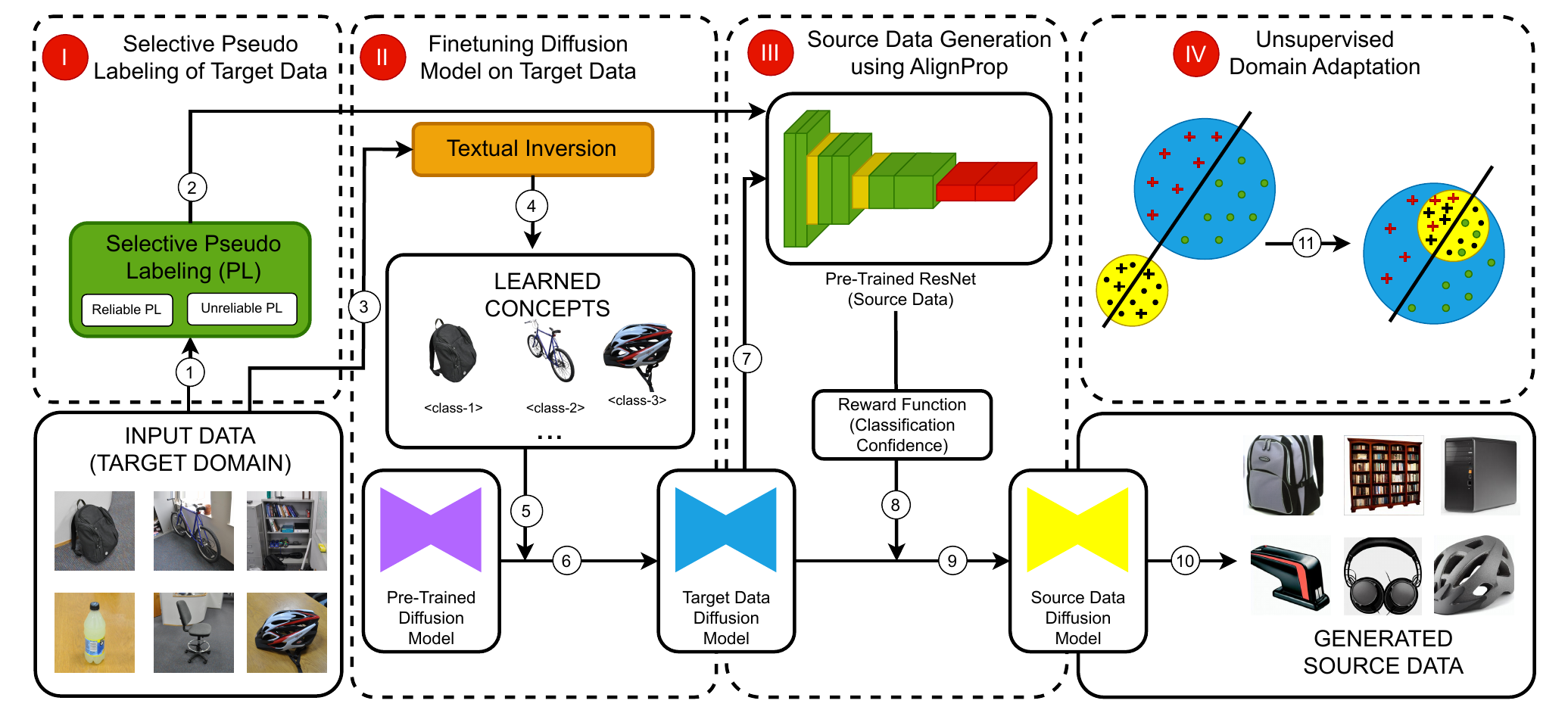}
    \caption{Overall training pipeline of the proposed DM-SFDA method. The training pipeline starts with selective pseudo labeling target data using the pre-trained source model. This is followed by fine-tuning a pre-trained text-to-image diffusion model on the target images using textual inversion \cite{textual_inversion}. Subsequently, the pre-trained source model is used to fine-tune this diffusion model using AlignProp \cite{alignprop} to generate Source Images. Finally, the finetuned diffusion models are used to generate intermediate domains between source and target domains to perform unsupervised domain adaptation.}
    \label{fig:overall}
\end{figure*}
\vspace{-4pt}
Most of the current state-of-the-art DA methods attain model adaptability by converging two disparate data distributions within a shared feature space, spanning both domains simultaneously~\cite{sfda-de}. One way of achieving this in a source-free manner is to use synthetically generated source data. However, generating synthetic source data that accurately represents the diversity and complexity of the source domain can be difficult. Furthermore, if the synthetic data is not of high quality, it might introduce noise and inconsistencies, negatively impacting the model’s performance on the target domain. Notably, recent advancements in Diffusion Generative Models (DGMs) \cite{diffusion, thermo} have demonstrated exceptional capabilities in producing diverse and high-quality images. Consequently, this paper aims to harness the generalizability of the state-of-the-art text-to-image diffusion models to the challenging task of SFDA.

To address the challenges of data privacy and diversity in the reconstruction process, we present an innovative framework named \textbf{D}iffusion \textbf{M}odels for \textbf{S}ource-\textbf{F}ree \textbf{D}omain \textbf{A}daptation (DM-SFDA). An overview of DM-SFDA is illustrated in Figure \ref{fig:overall}. The core idea of this approach is to use text-to-image diffusion models to generate images representative of the source domain based on the target domain and a pre-trained source network. Essentially, this involves fine-tuning a pre-trained text-to-image diffusion model to produce source samples that minimize the entropy for the pre-trained source model. The key contributions of our framework can be summarized as follows:

\begin{enumerate}
\item We propose a novel framework to enhance model performance in unseen domains while simultaneously addressing the challenges posed by limited access to source data and the increasing emphasis on data privacy.
 
\item Our novel framework harnesses the generalization capabilities of Diffusion Models to improve the data diversity and completeness within the reconstructed source data.

\item Through extensive qualitative and quantitative analyses of several traditional and state-of-the-art baselines and in-depth analysis, we demonstrate the effectiveness of the proposed pipeline.
\end{enumerate}

The remainder of the paper is organized as follows. In Section \ref{sec:rel}, we discuss related work in the areas of DA, SFDA, and DGMs. In Section \ref{sec:pre}, we revisit the preliminary concepts that form the basis of our proposed approach. In Section \ref{sec:method}, we discuss and formalize the problem definition and present our methodology. We present experimental results and analysis in Sections \ref{sec:res} and \ref{sec:ablation}. Finally, we discuss the limitations of our work in Section \ref{sec:challenge} and conclude with a brief summary in Section \ref{sec:conc}.
\vspace{-10pt}
\section{Related Work}
\label{sec:rel}
\subsection{Domain Adaptation}
DA has its roots in \cite{dafirst}, which focused on the role of good feature representations in their successful application for the task. The initial few works in DA adapted moment matching to align feature distributions between source and target domains~\cite{trans_feats, sel_bias, joint_adaptation_networks, easy_domain_adaptation, deep_domain_confusion}. Subsequent works used adversarial learning-based approaches  to tackle the problem of DA~\cite{da_backprop, classifier_discrepency, randomized_multi_adv_net, adv_da, dirt}. Apart from this, many other techniques like \cite{unsup_da, reg_deep_cluster, self_sup_da, multibranch_net} have been proposed to tackle the task of DA.

\subsection{Unsupervised Domain Adaptation}
Unsupervised Domain Adaptation (UDA) is a subtype of DA that aims to transfer knowledge from a labeled source domain to a different unlabeled target domain \cite{uda}. Existing mainstream UDA methods can be categorized into two main types of methods: those that align the source and
target domain distributions by designing specific metrics \cite{covi, fixbi, uda1, uda2}, and those that learn domain-invariant feature representations through adversarial learning \cite{gsde, uda4, uda5}. However, the success of most of these methods depends on a huge amount of source data which might not be available in most practical scenarios.

\subsection{Source-Free Domain Adaptation}
SFDA has been considered in the literature as a means of reducing reliance on source data. As described in \cite{sfdasurvey}, the existing SFDA research can generally be categorized into two approaches: data-centric and model-centric. Model-centric methods employ techniques such as self-training and self-attention, while data-centric methods include domain-based reconstruction and image-based information extraction.  Our proposed method follows the data-centric perspective to solve the SFDA task using source domain generation. 3C-GAN \cite{3c_gan} is a pioneering work in this area which uses a Generative Adversarial Network (GAN) to generate target-like images and simulatneously adapts the source pre-trained model. Some other works like SDDA \cite{sdda} and CPGA \cite{cpga} also solve the SFDA task using a similar data-generation approach. More recently, other works like AaD \cite{aad} and Co-learn \cite{colearn} have also shown state-of-the-art performance across various datasets and tasks.

\subsection{DGMs for Domain Adaptation}
Recently, there has been a significant shift in the landscape of generative modeling due to DGMs~\cite{diffusion, thermo}, demonstrating impressive capabilities in generating highly realistic text-conditioned images. DGMs have also seen a growing interest in the DA community with many recent works using DGMs as input augmentation techniques. A recent example is a text-to-image diffusion model, employed by \cite{datum} to generate target domain images using source domain labels, thereby demonstrating the efficacy of diffusion models in One-Shot Unsupervised Domain Adaptation (OSUDA). DGMs, when trained on multiple source domains, have also been instrumental in guiding approximate inference in target domains, as reported by \cite{diffusion_priors}. In our work, we leverage a recently introduced fine-tuning strategy for diffusion models called AlignProp~\cite{alignprop}, to fine-tune the diffusion models using the output probability of the source model as an objective function.

\section{Preliminaries}
\label{sec:pre}
\subsection{Conditional Diffusion Probabilistic Models}
Conditional Denoising Diffusion Probabilistic Model (DDPM) forms the backbone of our source data reconstruction pipeline. This model represent a distribution over data $x_0$, conditioned on a contextual input labeled $c$. This distribution arises from a sequential denoising process, which aims to reverse a Markovian forward process denoted as $q(x_t | x_{t-1})$. This forward process progressively introduces a Gaussian noise to the data. Subsequently, a forward process posterior mean predictor $\mu_\theta(x_t,t,c)$ is then trained to reverse the forward process for all $t \in \{ 0, 1, 2, ..., T \}$. The training process entails maximizing a variational lower bound on the model log-likelihood with an objective function defined as follows:

\begin{equation}
\label{eq:ddpm}
    \mathcal{L}_{DDPM}(\theta) = \mathbb{E} [|| \hat{\mu}(x_t, x_0) - \mu_\theta(x_t,t,c)||^2]
\end{equation}
where $\hat{\mu}$ is a weighted average of $x_0$ and $x_t$.

The sampling process of a diffusion model starts with a sample from the Gaussian distribution $x_T \sim \mathcal{N}(0,1)$ which is subsequently denoised using the reverse process $p_\theta(x_{t-1} | x_t, c)$ to product a trajectory $\{ x_t, x_{T-1}, ..., x_0 \}$ ending with a sample $x_0$. Here, the sampler use an isotropic Gaussian reverse process with a fixed time-dependent variance:

\begin{equation}
\label{eq:sampler}
    p_\theta(x_{t-1} | x_t, c) = \mathcal{N}(x_{t-1} | \mu_\theta(x_t,t,c), \sigma_t^2\textbf{I})
\end{equation}

\subsection{Markov Decision Process}
A Markov decision process (MDP) serves as a structured representation of problems involving a series of interconnected decisions. It is characterized by a set of components denoted as $(\mathcal{S}, \mathcal{A}, \rho_0, P, R)$, where $\mathcal{S}$ signifies the collection of possible states, $\mathcal{A}$ stands for the available actions, $\rho_0$ signifies the initial state distribution, $P$ represents the transition pattern, and $R$ defines the reward mechanism. At each discrete time step denoted as $t$, the agent observes a specific state represented as $s_t$ from the state space $S$ takes an action labeled as $a_t$ from the action space $\mathcal{A}$ and in response receives a reward labeled as $R(s_t, a_t)$, subsequently transitioning to a novel state $s_{t+1}$ drawn from the distribution $P(\cdot | s_t, a_t)$. The agent's decision-making is guided by a policy $\pi(a | s)$ that dictates actions based on states. During the agent's engagement with the MDP, it generates trajectories, which are sequences comprising both states and actions, conventionally presented as $\tau = (s_0, a_0, s_1, a_1, . . . , s_T, a_T )$.

\label{sec:method}

\section{Proposed Method}
\label{sec:method}
\subsection{Notations and Problem Definition}
Following the notations used in \cite{sfdasurvey}, in this paper, we represent a domain as $\mathcal{D}$. Each domain consists of a dataset $\phi$ and an associated label set $\mathbf{L}$. A dataset comprises of an instance set $\mathcal{X} = { x_i}_{i=1}^{n}$, derived from a $d$-dimensional marginal distribution $\mathcal{P} (\mathcal{X})$, and a label set $\mathcal{Y} = { y_i}_{i=1}^{c}$ where $n$ represents the total number of samples and $c$ represents the total number of classes.

The SFDA scenario involves two stages: pre-training and adaptation. During pre-training, a model $\mathcal{M}$ is trained on labeled data from the source domain $\mathcal{D}^S = \{\{\mathcal{X}^s,\mathcal{P} (\mathcal{X}^s), d^s\},\mathbf{L}^s\}$. Subsequently, the goal of the adaptation phase is to adapt the pre-trained source model to the unlabeled target data $\phi^t = {{\mathcal{X}^t,\mathcal{P} (\mathcal{X}^t), d^t}}$. The proposed approach assumes a closed form, implying that the label spaces of the source and target domains are identical.

\subsection{Method Overview}
An overview of the proposed method DM-SFDA for source free domain adaptation using text-to-image diffusion models can be seen in Figure~\ref{fig:overall}. The key idea behind the approach is to leverage the generalizability of the state-of-the-art text-to-image diffusion models to tackle the task of SFDA. The training pipeline contains the following four phases: I) Selective Pseudo Labeling Target Data II) Fine-tuning Diffusion Model on Target Data, III) Source Data Generation using AlignProp, IV) Unsupervised Domain Adaptation. The first phase involves selective pseudo labeling target data using the pre-trained source model. The second phase involves finetuning a pre-trained diffusion model to learn concepts in the target domain using Textual Inversion \cite{textual_inversion}. Subsequently, the pre-trained source model is used to fine-tune this diffusion model using AlignProp \cite{alignprop} to generate Source Images. Finally, a diffusion model-based domain mixup strategy is used to perform unsupervised domain adaptation. Each of these phases are described in detail in the following subsections.

\subsection{Phase I: Selective Pseudo Labeling Target Data}
\label{sec:caption}
The proposed DM-SFDA pipeline requires labeled target data to generate data from the source domain. Therefore, the initial phase of our proposed pipeline addresses the challenge of unlabeled target data by selecting reliable labels for the target samples using a selective pseudo-labeling strategy. This approach is akin to the one proposed in \cite{csfda}. As shown in \cite{conf_da}, prediction confidence and difference in entropy can be reliable measures of pseudo-label accuracy and estimate different types of domain shifts. Therefore, prediction confidence and average entropy are used as metrics to assess label reliability. A binary reliability score ($r^i$) is assigned to each sample in the target data, determined by their prediction confidence and prediction uncertainty ($g_u^i$), as illustrated below:

\begin{equation}
    \begin{array}{c} 
        g_u^i = std\{conf(\mathcal{M}(x^t))\} \\ \\

        \mathcal{T}_c = \frac{1}{B} \sum_{i=1}^B conf(\mathcal{M}(x^t)) \\ \\

        \mathcal{T}_u = \frac{1}{B} \sum_{i=1}^B g_u^i \\ \\
        
        r^i = \begin{cases}
            1, & \textrm{if } conf(\mathcal{M}(x^t)) \geq \mathcal{T}_c \textrm{ and } g_u^i \leq \mathcal{T}_u\\
            0, & otherwise
        \end{cases}
    \end{array}
\end{equation}

Here, $\mathcal{T}_c$ and $\mathcal{T}_u$ represent the selection thresholds for confidence and uncertainity. Taking the average as a threshold eliminates the requirement of per-dataset hyper-parameter tuning and makes the selection process highly adaptive. Furthermore, aleatoric uncertainty \cite{aelotrophic} is used since it better addresses the concern of domain shift.

\subsection{Phase II: Diffusion Model Finetuning on Target Data}
The second phase of the proposed DM-SFDA pipeline involves fine-tuning a text-to-image diffusion model on the target data using LoRA. However, the lack of class labels poses a challenge as there are no textual cues to guide the diffusion process. To address this, we employ a recently introduced fine-tuning strategy called Textual Inversion \cite{textual_inversion}. Using images from the target domain, Textual Inversion learns to represent objects in the images through new "words" in the embedding space of a pre-trained text-to-image model. As illustrated in Figure \ref{fig:overall}, we assign a placeholder string "<class-\{idx\}>" for the newly learned concepts, using class indices from the selective pseudo labeling done in Phase I. These new identifiers are then used as textual cues to guide the diffusion process in subsequent phases of the pipeline.

\subsection{Phase III: Source Data Generation using AlignProp}
In the third phase, we use the method proposed in \cite{alignprop} to further fine-tune our diffusion model to generate source-like images. This finetuning is done by transforming the denoising process of a diffusion model into a differentiable recurrent policy. The iterative denoising process is mapped to the following single-step MDP:

\begin{equation}
    \begin{array}{c} 
    \mathcal{S} \triangleq \{(x_T , c), x_T \sim \mathcal{N} (0, 1)\} \\  \\ 
    \mathcal{A} \triangleq \{ x_0 : x_0 \sim \pi_\theta(\cdot | x_T , c), x_T \sim \mathcal{N} (0, 1) \} \\ \\
    R_\phi(x_0), x_0 \in \mathcal{A} \\ \\
    \end{array}
\end{equation}

Here, $R_\phi$ is the reward function which is dependent on the generated images. In our case, we define the reward function to use the information in the pre-trained source model to guide the diffusion process to generate source-like images. Inspired by the loss function used in DAFL \cite{DAFL}, our reward function consists of the following three components to extract maximum information from the source model:

\begin{itemize}
    \item \textbf{Confidence Reward}: The confidence reward function $R_{conf}$ makes sure that higher confidence predictions are assigned a higher reward. 
    \begin{equation}
        R_{conf} = \frac{1}{B} \sum_{i=1}^B conf(\mathcal{M}(x^t))
    \end{equation}
    \item \textbf{One-Hot Reward}: As described in \cite{DAFL}, the outputs of the source-model should be similar to the training data if the input follows the training distribution. Therefore, the one-hot reward function $R_{OH}$ assigns higer rewards to the samples that generate one-hot like predictions.
    \begin{equation}
        R_{OH} = 1 - \frac{1}{B} \sum_i \mathcal{H}_{cross}(\mathcal{M}(x^t), y^t)
    \end{equation}
    \item \textbf{Batch Norm Statistics (BNS) Reward}: As described in \cite{bns}, to effectively match the low level and high level feature maps of the generated images, we can match the Batch Norm Statistics of the pre-trained neural network with the generated images. The BNS reward function ensures that a higer reward is given to the samples which have similar BNS as that of the model.
    \begin{equation}
        R_{BNS} = - ( || \sum_l \mu_l(x^t) - BN_l(running\_mean) ||_2 + || \sum_l \sigma_l(x^t) - BN_l(running\_variance) ||_2 )
    \end{equation}
\end{itemize}

The final reward function used for finetuning the diffusion model is defined as:

\begin{equation}
    R = \lambda_A R_{conf} + \lambda_B R_{OH} + \lambda_C R_{BNS}
\end{equation}

Using the above state, action and reward functions, and the class prompts $\mathcal{P}$, the parameters of the diffusion model are updated using gradient descent on the following loss function:

\begin{equation}
    \mathcal{L}_{align}(\theta; \mathcal{P}) = - \frac{1}{|\mathcal{P}|} \sum_{c^i \in \mathcal{P}} R_\phi(\pi_\theta(x^t , c^i))
\end{equation}

The training is done using backpropagation through time on the recurrent policy $\pi$. Furthermore, as described in \cite{alignprop}, in order to reduce the memory overload, the LoRA weights are finetuned along with using a truncated backpropagation through time (TBTT) instead of a full backpropagation through time (BPTT). After the completion of the AlignProp fine-tuning, the diffusion model produces source images by utilizing the "<class-\{idx\}>" placeholder as guiding prompt.

\begin{figure}
    \centering
    \includegraphics[scale=0.4]{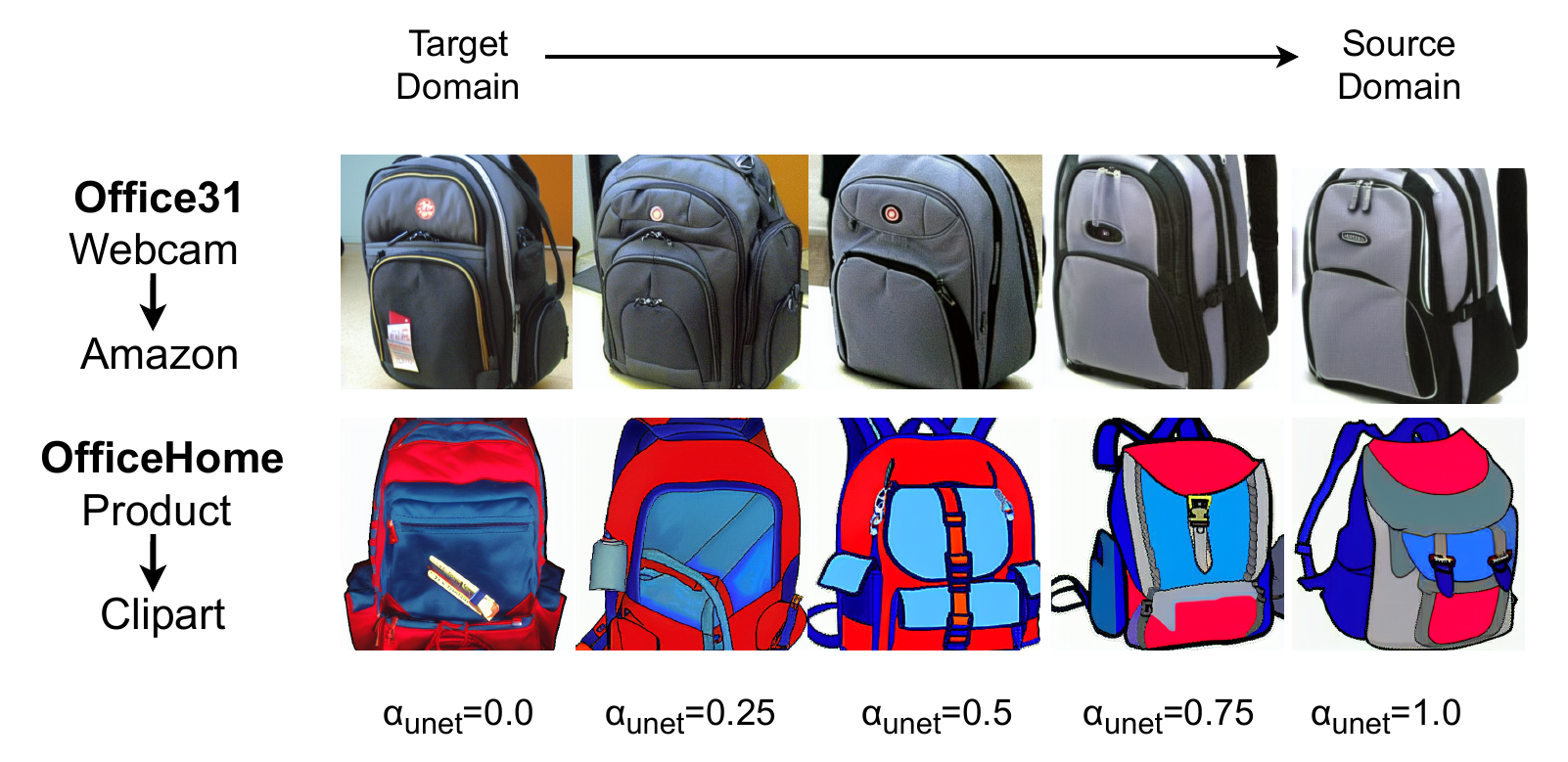}
    \caption{Visualization of the diffusion-based domain mixup for Unsupervised Domain Adaptation.}
    \label{fig:mixup}
\end{figure}

\subsection{Phase IV: Unsupervised Domain Adaptation}
The fourth phase starts by labeling the generated source domain images using the pre-trained source model. Once we have the reconstructed and labeled source domain data, we have effectively converted the initial problem of SFDA to a standard Unsupervised Domain Adaptation (UDA) problem. Since, all of the diffusion model finetuning happens via Low-Rank Adapters, we can use the finetuned weight patches to effectively compensate the domain discrepancy between the source and target domain. This is done by generating multiple intermediate augmented domains by altering the scaling parameter $\alpha_{unet}$ while applying the LoRA patch to the pretrained model. Figure \ref{fig:mixup}, shows the visualization of the diffusion model-based inter-domain mixup. Subsequently, we use the approach proposed in \cite{fixbi} to train two complementary models on each of these intermediate domains that teach each other to bridge the domain gap by using a confidence-based learning where one model teaches the other model using the positive pseudo-labels or teach itself using the negative pseudo-labels. Through the confidence-based learning approach, the two models with different characteristics gradually get closer to the target domain. Furthermore, a consistency regularization is used to ensure a stable convergence of training both models. 

\begin{table*}[h]
    \setlength{\tabcolsep}{2pt}
    \def\arraystretch{1.2}
    \tiny
    \centering
    \begin{tabular}{c|c|cccccc|c}
        \toprule
        Method & \specialcell{Source\\Free} & $A \rightarrow D$ & $A \rightarrow W$ & $D \rightarrow A$ & $D \rightarrow W$ & $W \rightarrow A$ & $W \rightarrow D$ & Avg. \\
        \midrule
        ResNet-50 \cite{resnet} & \xmark & 68.9 & 68.4 & 62.5 & 96.7 & 60.7 & 99.3 & 76.1 \\
        \midrule

        MCC \cite{mcc} & \xmark & 95.6 & 95.4 & 72.6 & 98.6 & 73.9 & 100.0 & 89.4 \\
        GSDA \cite{gsda} & \xmark & 94.8 & 95.7 & 73.5 & 99.1 & 74.9 & 100.0 & 89.7 \\

        SRDC \cite{srdc} & \xmark & 95.8 & 95.7 & 76.7 & 99.2 & 77.1 & 100.0 & 90.8 \\
        FixBi \cite{fixbi} & \xmark & 95.0 & 96.1 & 78.7 & 99.3 & 79.4 & 100.0 & 91.4 \\
        CoVi \cite{covi} & \xmark & 98.0 & 97.6 & 77.5 & 99.3 & 78.4 & 100.0 & 91.8 \\
        ICON \cite{icon} & \xmark & 97.0 & 93.3 & 79.4 & 99.2 & 78.3 & 100.0 & 91.2 \\
        \midrule
        SHOT \cite{shot} & \cmark & 94.0 & 90.1 & 74.7 & 98.4 & 74.3 & 99.9 & 88.6 \\
        3C-GAN \cite{3c_gan} & \cmark & 92.7 & 93.7 & 75.3 & 98.5 & 77.8 & 99.8 & 89.6 \\
        NRC \cite{nrc} & \cmark & 96.0 & 90.8 & 75.3 & 99.0 & 75.0 & 100.0 & 89.4 \\

        NRC++ \cite{nrcpp} & \cmark & 95.9 & 91.2 & 75.5 & 99.1 & 75.0 & 100.0 & 89.5 \\

        AaD \cite{aad} & \cmark & 96.4 & 92.1 & 75.0 & 99.1 & 76.5 & 100.0 & 89.9 \\
        AaD w/ Co-learn \cite{colearn} & \cmark & 97.6 & 98.7 & 82.1 & 99.3 & 80.1 & 100.0 & 93.0 \\
        \midrule
        \rowcolor{Gray}
        DM-SFDA & \cmark & \textbf{97.7} & \textbf{99.0} & \textbf{82.7} & \textbf{99.3} & \textbf{83.5} & \textbf{100.0} & \textbf{93.7} \\
        \bottomrule
    \end{tabular}
    \caption{Classification accuracy (\%) under UDA and SFDA settings on Office-31 \cite{office31} dataset for source-free domain adaptation (ResNet-50). Best results under SFDA setting are shown in bold font.}
    \label{tab:office31}
\end{table*}

\begin{table*}[h]
    \def\arraystretch{1.2}
    \tiny
    \centering
    \begin{tabular}{c @{\hskip 0.05in} | @{\hskip 0.05in} c @{\hskip 0.05in} | @{\hskip 0.05in}ccc@{\hskip 0.05in} | @{\hskip 0.05in}ccc@{\hskip 0.05in} | @{\hskip 0.05in}ccc@{\hskip 0.05in} | @{\hskip 0.05in}ccc@{\hskip 0.05in} | @{\hskip 0.05in}c}
        \toprule
        Method & Source & \multicolumn{3}{c|@{\hskip 0.05in}}{Ar $\rightarrow$} & \multicolumn{3}{c|@{\hskip 0.05in}}{Cl $\rightarrow$} & \multicolumn{3}{c|@{\hskip 0.05in}}{Pr $\rightarrow$} & \multicolumn{3}{c|@{\hskip 0.05in}}{Rw $\rightarrow$} & Avg. \\
         & Free & Cl & Pr & Rw & Ar & Pr & Rw & Ar & Cl & Rw & Ar & Cl & Pr & \\
        \midrule
        ResNet-50 \cite{resnet} & \xmark & 34.9 & 50.0 & 58.0 & 37.4 & 41.9 & 46.2 & 38.5 & 31.2 & 60.4 & 53.9 & 41.2 & 59.9 & 46.1 \\
        \midrule

        MCC \cite{mcc} & \xmark & 88.1 & 80.3 & 80.5 & 71.5 & 90.1 & 93.2 & 85.0 & 71.6 & 89.4 & 73.8 & 85.0 & 36.9 & 78.8\\
        
        GSDA \cite{gsda} & \xmark & 61.3 & 76.1 & 79.4 & 65.4 & 73.3 & 74.3 & 65.0 & 53.2 & 80.0 & 72.2 & 60.6 & 83.1 & 70.3 \\
        SRDC \cite{srdc} & \xmark & 52.3 & 76.3 & 81.0 & 69.5 & 76.2 & 78.0 & 68.7 & 53.8 & 81.7 & 76.3 & 57.1 & 85.0 & 71.3 \\
        FixBi \cite{fixbi} & \xmark & 58.1 & 77.3 & 80.4 & 67.7 & 79.5 & 78.1 & 65.8 & 57.9 & 81.7 & 76.4 & 62.9 & 86.7 & 72.7 \\

        CoVi \cite{covi} & \xmark & 58.5 & 78.1 & 80.0 & 68.1 & 80.0 & 77.0 & 66.4 & 60.2 & 82.1 & 76.6 & 63.6 & 86.5 & 73.1 \\
        ICON \cite{icon} & \xmark & 63.3 & 81.3 & 84.5 & 70.3 & 82.1 & 81.0 & 70.3 & 61.8 & 83.7 & 75.6 & 68.6 & 87.3 & 75.8 \\
        
        \midrule
        SHOT \cite{shot} & \cmark & 57.1 & 78.1 & 81.5 & 68.0 & 78.2 & 78.1 & 67.4 & 54.9 & 82.2 & 73.3 & 58.8 & 84.3 & 71.8 \\
        NRC \cite{nrc} & \cmark & 57.7 & 80.3 & 82.0 & 68.1 & 79.8 & 78.6 & 65.3 & 56.4 & 83.0 & 71.0 & 58.6 & 85.6 & 72.2 \\
        NRC++ \cite{nrcpp} & \cmark & 57.8 & 80.4 & 81.6 & 69.0 & 80.3 & 79.5 & 65.6 & 57.0 & 83.2 & 72.3 & 59.6 & 85.7 & 72.5 \\
        AaD \cite{aad} & \cmark & 59.3 & 79.3 & 82.1 & 68.9 & 79.8 & 79.5 & 67.2 & 57.4 & 83.1 & 72.1 & 58.5 & 85.4 & 72.7 \\
        AaD w/ Co-learn \cite{colearn} & \cmark & 65.1 & 86.0 & \textbf{87.0} & \textbf{76.8} & \textbf{86.3} & 86.5 & \textbf{74.4} & 66.1 & 87.7 & \textbf{77.9} & 66.1 & 88.4 & 79.0 \\
        \midrule
        \rowcolor{Gray}
        DM-SFDA & \cmark & \textbf{68.5} & \textbf{89.6} & 83.3 & 70.0 & 85.8 & \textbf{87.4} & 71.3 & \textbf{69.6} & \textbf{88.2} & 77.8 & \textbf{68.5} & \textbf{88.7} & \textbf{79.5}\\
        \bottomrule
    \end{tabular}
    \caption{Classification performance (\%) under UDA and SFDA settings on Office-Home dataset \cite{office_home} (ResNet-50 backbone). We report Top-1 accuracy on 12 domain shifts (→) and take the average (Avg.) over them.}
    \label{tab:office_home}
\end{table*}

\begin{table*}[h]
    \setlength{\tabcolsep}{2pt}
    \def\arraystretch{1.2}
    \centering
    \resizebox{0.8\textwidth}{!}{    
    \begin{tabular}{c|c|cccccccccccc|c}
        \toprule
        Method & Source-Free & \texttt{plane} & \texttt{bike} & \texttt{bus} & \texttt{car} & \texttt{horse} & \texttt{knife} & \texttt{mcycle} & \texttt{person} & \texttt{plant} & \texttt{sktbrd} & \texttt{train} & \texttt{truck} & Avg. \\
        \midrule
        ResNet-101 \cite{resnet} & \xmark & 55.1 & 53.3 & 61.9 & 59.1 & 80.6 & 17.9 & 79.7 & 31.2 & 81.0 & 26.5 & 73.5 & 8.5 & 52.4  \\
        \midrule
        MCC \cite{mcc} & \xmark & 88.7 & 80.3 & 80.5 & 71.5 & 90.1 & 93.2 & 85.0 & 71.6 & 89.4 & 73.8 & 85.0 & 36.9 & 78.8 \\
        GSDA \cite{gsda} & \xmark & 93.1 & 67.8 & 83.1 & 83.4 & 94.7 & 93.4 & 93.4 & 79.5 & 93.0 & 88.8 & 83.4 & 36.7 & 81.5 \\
        FixBi \cite{fixbi} & \xmark & 96.1 & 87.8 & 90.5 & 90.3 & 96.8 & 95.3 & 92.8 & 88.7 & 97.2 & 94.2 & 90.9 & 25.7 & 87.2\\
        CoVi \cite{covi} & \xmark & 96.8 & 85.6 & 88.9 & 88.6 & 97.8 & 93.4 & 91.9 & 87.6 & 96.0 & 93.8 & 93.6 & 48.1 & 88.5 \\
        ICON \cite{icon} & \xmark & 96.6 & 91.8 & 88.0 & 82.0 & 96.8 & 93.3 & 90.0 & 81.0 & 95.2 & 93.6 & 91.0 & 49.5 & 87.4 \\
        \midrule
        SHOT \cite{shot} & \cmark & 94.3 & 88.5 & 80.1 & 57.3 & 93.1 & 94.9 & 80.7 & 80.3 & 91.5 & 89.1 & 86.3 & 58.2 & 82.9 \\
        3C-GAN \cite{3c_gan} & \cmark & 94.8 & 73.4 & 68.8 & 74.8 & 93.1 & 95.4 & 88.6 & 84.7 & 89.1 & 84.7 & 83.5 & 48.1 & 81.6 \\
        NRC \cite{nrc} & \cmark & 96.8 & 91.3 & 82.4 & 62.4 & 96.2 & 95.9 & 86.1 & 80.6 & 94.8 & 94.1 & 90.4 & 59.7 & 85.9 \\
        NRC++ \cite{nrcpp} & \cmark & 96.8 & 91.9 & 88.2 & 82.8 & 97.1 & \textbf{96.2} & 90.0 & 81.1 & 95.2 & 93.8 & 91.1 & 49.6 & 87.8 \\
        AaD \cite{aad} & \cmark & 97.4 & \textbf{90.5} & 80.8 & 76.2 & \textbf{97.3} & 96.1 & 89.8 & 82.9 & 95.5 & 93.0 & 92.0 & \textbf{64.7} & 88.0 \\
        AaD w/ Co-learn \cite{colearn} & \cmark & 97.6 & 90.2 & 85.0 & 83.1 & 97.1 & 92.1 & 84.9 & 96.8 & \textbf{96.8} & \textbf{95.1} & 92.2 & 56.8 & \textbf{89.1} \\
        \midrule
        \rowcolor{Gray}
        DM-SFDA & \cmark & \textbf{98.1} & 89.8 & \textbf{90.6} & \textbf{90.5} & 96.8 & 95.2 & \textbf{92.2} & 93.4 & 97.8 & 94.4 & \textbf{92.4} & 48.8 & 86.3 \\
        \bottomrule
    \end{tabular}
    }
    \caption{Per-class accuracy and mean accuracy (\%) on VisDA-2017 \cite{visda2017} dateset for source-free domain adaptation (ResNet-101). Best results under SFDA setting are shown in bold font.
}
    \label{tab:visda}
\end{table*}

\section{Experiments and Results}
\label{sec:res}
\subsection{Datasets}
    \begin{itemize}
        \item \textbf{Office-31:} Office-31~\cite{office31} is a benchmark image classification dataset that consists of a limited set of images distributed across 31 categories spanning three domains: Amazon (2,817 images), DSLR (498 images), and Webcam (795 images).
        \item \textbf{Office-Home:} Office-Home~\cite{office_home}, on the other hand, comprises a more extensive dataset with a total of 15.5K images from 65 classes, gathered from 4 distinct image domains: Artistic, Clipart, Product, and Real-world. Our analysis includes 12 transfer tasks for this dataset.
        \item \textbf{VisDA:} VisDA~\cite{visda2017} encompasses two distinct domains: synthetic and real, each comprising 12 classes. The synthetic domain holds around 150K computer-generated 3D images with various poses, while the corresponding real domain includes approximately 55K images captured from the real world.
    \end{itemize}

\subsection{Experimental Setup}
We implement our approach using PyTorch and use ResNet-50 \cite{resnet} as the backbone network for the Office-31~\cite{office31} and Office-Home~\cite{office_home} datasets and Resnet-101 for the VisDA~\cite{visda2017} dataset. All the experiments for our proposed approach were conducted on a Nvidia A100 GPU. The other details for specific parts of our pipeline are specified below:

\begin{itemize}
    \item \textbf{Diffusion Model Fine-tuning:} We use Stable Diffusion v1.4 \cite{diffusion2} as the base model for all experiments. The finetuning of the diffusion models was done in the Accelerate~\cite{accelerate} environment. Memory-efficient attention was enabled using xFormers~\cite{xFormers2022} for all the experiments performed. Furthermore, we used the Low-Rank Adaptation (LORA) of the Stable Diffusion Pipeline.
    \item \textbf{AlignProp:} In the context of the AlignProp experiments, we employed the Low-Rank Adaptation (LORA) technique within the framework of the Stable Diffusion Pipeline. The training was done for a batch size of 4, and 100 batches were sampled per step. The training procedure encompassed 100 steps, with each step incorporating two distinct phases: a sampling phase and 10 consecutive inner training epochs dedicated to training the model on the sampled data from the previous phase.
    \item \textbf{Unupervised Domain Adaptation:} In the unsupervised domain adaptation phase of our pipeline, we use the experimental setup proposed in \cite{fixbi}. For the UDA approach, we use the generated source data labeled using the pre-trained model and all unlabeled target data. We use minibatch stochastic gradient descent (SGD) with a momentum of 0.9, an initial learning rate of 0.001, and a weight decay of 0.005. We follow the same learning rate schedule as in \cite{scheduler}.
\end{itemize}


\subsection{Results}
\subsubsection{Results on Office-31}
We present the outcomes for the Office-31 dataset in Table~\ref{tab:office31}. By employing the source data produced through our proposed pipeline as input along with the diffusion-model based mixup strategy, our approach is able to match/outperform existing state-of-the-art SFDA approaches for all the tasks, achieving an average accuracy of \textbf{93.7\%}, which is 0.7\% higher then the current state-of-the-art.

\subsubsection{Results on Office-Home}
The results for the Office-Home dataset \cite{office_home} are presented in Table \ref{tab:office_home}. By integrating the source data generated through our proposed pipeline into the proposed UDA mixup methodology, we have successfully surpassed existing methods, achieving an average accuracy of \textbf{79.5\%}.

\subsubsection{Results on VisDA17}
We summarize the results for the VisDA dataset \cite{visda2017} in Table~\ref{tab:visda}. Our proposed framework is able to outperform the source-free baselines across many classes like plane, bus and car by \textbf{0.5\%},  \textbf{2.1\%} and \textbf{7.4\%} respectively. The accuracy of our proposed method surpass the approaches where source data is available, thereby showing the effectiveness of our data generation and domain mixup pipeline.

\section{Ablation Study}
\label{sec:ablation}
In this section, we perform an ablation study to understand the contribution of each component in the proposed pipeline to the overall performance. 

\subsection{Selective Pseudo Labeling Target Data}
Correct pseudo labeling of target data samples is essential for the success of the proposed approach, as it significantly influences the fine-tuning process of the diffusion model. To gauge its impact, we assess the model's performance without employing selective pseudo labeling. This will help assess the importance of initial pseudo labels in guiding the subsequent fine-tuning and adaptation phases. As shown in Table \ref{tab:spl}, a significant drop in performance is observed for the proposed pipeline in the absence of the selective pseudo labeling. The primary reason for this is the inaccuracies in pseudo label assignment during the initial phase that adversely affect all subsequent phases of the pipeline, including data generation and UDA. 

\begin{table}[h]
    \setlength{\tabcolsep}{2pt}
    \def\arraystretch{1.2}
    \tiny
    \centering
    \begin{tabular}{c|cccccc|c}
        \toprule
        \specialcell{Selective \\ Pseudo Labeling} & $A \rightarrow D$ & $A \rightarrow W$ & $D \rightarrow A$ & $D \rightarrow W$ & $W \rightarrow A$ & $W \rightarrow D$ & Avg. \\
        \midrule
        \xmark & 67.8 & 68.3 & 60.1 & 95.4 & 60.5 & 98.7 & 75.1 \\
        \cmark & 97.7 & 99.0 & 82.7 & 99.3 & 83.5 & 100.0 & 93.7 \\
        \bottomrule
    \end{tabular}
    \vspace{2mm}    
    \caption{Ablation study to investigate effects of selective pseudo-labeling.}
    \label{tab:spl}
\end{table}
\vspace{-17pt}

\subsection{Unsupervised Domain Adaptation}
In order to test the efficacy of our Diffusion-Model based Unsupervised Domain Adaptation approach, we compare the downstream performance of our proposed approach with the existing off-the-shelf UDA approaches. Since we chose a ResNet backbone as the pre-trained source model, we experiment with the current state-of-the-art ResNet-based UDA approaches. As shown in Table \ref{tab:uda}, our proposed diffusion model-based mixup approach is able to significantly outperform all other UDA approaches. This shows that our mixup approach is able to generate much better intermediate domain to better bridge the huge domain gap.

\begin{table}[h]
    \setlength{\tabcolsep}{2pt}
    \def\arraystretch{1.2}
    \tiny
    \centering
    \begin{tabular}{c|cccccc|c}
        \toprule
        \specialcell{Method} & $A \rightarrow D$ & $A \rightarrow W$ & $D \rightarrow A$ & $D \rightarrow W$ & $W \rightarrow A$ & $W \rightarrow D$ & Avg. \\
        \midrule
        DM-SFDA (FixBi \cite{covi}) & 93.0 & 93.5 & 77.4 & 98.7 & 78.3 & 99.7 & 90.1 \\
        DM-SFDA (CoVi \cite{fixbi}) & 93.6 & 94.0 & 77.0 & 99.0 & 78.0  & 99.9  & 90.2\\
        DM-SFDA (ICON \cite{icon}) & 95.2 & 92.9 & 78.6 & 99.2 & 78.2 & 100.0 & 90.7 \\
        DM-SFDA & 97.7 & 99.0 & 82.7 & 99.3 & 83.5 & 100.0 & 93.7 \\
        \bottomrule
    \end{tabular}
    \vspace{2mm}
    \caption{Ablation results of prevalent of-the-shelf UDA methods applied to our generated source and target images as compared to our proposed diffusion model-based mixup approach.}
    \label{tab:uda}
\end{table}
\vspace{-20pt}
\section{Challenges and Limitations}
\label{sec:challenge}
\textbf{Computational Resources}: Training and running the proposed pipeline to generate source-like images is computationally intensive and time-consuming, requiring significant computational resources. This may limit the practicality of the proposed approach for researchers or practitioners with limited access to such resources.

\textbf{Scalability to Different Domains}: The effectiveness of the proposed method in different application domains has yet to be fully explored. Some domains might present unique challenges that are not adequately addressed by the current model, such as highly structured where slight inaccuracies in generated data could lead to significant performance drops.

\section{Conclusion}
\label{sec:conc}
In this work, we presented a novel approach to tackle the problem of SFDA by using a text-to-image diffusion model to generate source images with high confidence predictions by the pre-trained models. We conduct extensive experiments on multiple domain adaptation benchmarks. Compared with recent data-based domain adaptation methods, our model achieves the best or comparable performance in the absence of source data, thereby proving the efficacy of our proposed approach.

\bibliography{neurips_2024}
\bibliographystyle{plainnat}

\end{document}